\documentclass{article}
\usepackage{spconf,amsmath,graphicx,url}
\usepackage{enumerate}
\title{English Broadcast News Speech Recognition by Humans and Machines}

\twoauthors
 {\shortstack{Samuel Thomas, Masayuki Suzuki$^{*}$, \\
   Yinghui Huang, Gakuto Kurata$^{*}$, Zoltan Tuske, \\
   George Saon, Brian Kingsbury, Michael Picheny}}
	{$^{*}$IBM Research AI, Tokyo, Japan\\
	IBM Research AI, Yorktown Heights, USA}
  {\shortstack{Tom Dibert, Alice Kaiser-Schatzlein, \\Bern Samko}}
	{Appen, Sydney, Australia}

\begin{document}
\ninept
\maketitle
\begin{abstract}
With recent advances in deep learning, considerable attention has been given to achieving automatic speech recognition 
performance close to human performance on tasks like conversational telephone speech (CTS) recognition.
In this paper we evaluate the usefulness of these proposed techniques on broadcast news (BN), 
a similar challenging task. We also perform a set of recognition measurements to understand 
how close the achieved automatic speech recognition results are to human performance on this task. 
On two publicly available BN test sets,  DEV04F and RT04, our speech recognition system 
using LSTM and residual network based acoustic models with a combination of n-gram and neural network 
language models performs at 6.5\% and 5.9\% word error rate. By achieving new performance milestones on 
these test sets, our experiments show that techniques developed on other related tasks, like CTS, can be transferred 
to achieve similar performance. In contrast, the best measured human recognition performance on these 
test sets is much lower, at  3.6\% and 2.8\% respectively,  indicating that there is still room for new techniques 
and improvements in this space, to reach human performance levels.
\end{abstract}
\begin{keywords}
Broadcast News, Automatic Speech Recognition, Deep neural networks.
\end{keywords}

\section{Introduction}
\label{sec:intro}

Prior to the recent ubiquitous deployment of automatic speech recognition technology for various device user interfaces, two key domains of interest for application of automatic speech recognition technology were conversational telephone speech (CTS)
and broadcast news (BN). Interest in these domains was primarily fueled by various DARPA programs \cite{olive2011handbook}.
More recently, by employing various deep learning techniques, performance of speech recognition systems on the CTS task 
is getting close to human parity.  Several sites have made significant progress to lower the WER 
to within the 5\%-10\% range on the Switchboard-CallHome subsets of the Hub5 2000 
evaluation \cite{saon2017english,kurata2017language,xiong2018microsoft,han2017capio}. Given the progress 
on conversational telephone speech, we focus on the other closely related broadcast news recognition task that received 
similar attention within the DARPA EARS program. One of the key objectives of this study is to understand how deep 
learning based techniques developed on CTS generalize to the BN task.

In the BN domain, speech recognition systems need to deal with wide-band signals collected over a wide variety 
of speakers with different speaking styles, in various background noise conditions, and speaking on a wide variety 
of news topics. Most of the speech is well articulated and is formed similarly to written English. 
In contrast, CTS is spontaneous speech recorded over a telephone channel that introduces additional artifacts 
in addition to numerous speaking styles. Conversational speech is  interspersed with portions of overlapping 
speech, interruptions, restarts and back-channel confirmations between participants. In terms of the 
amount of training data available from the DARPA EARS program for training systems on CTS and BN, there 
are a few significant differences as well. The CTS acoustic training corpus consists of approximately 2000 hours of speech 
with human transcriptions \cite{saon2017english}. On the other hand, for the BN task, only about 140 hours of data is carefully transcribed. 
The remaining $\sim$9000 hours of available speech are TV shows with closed captions. In other words, models being 
developed for BN typically use lightly supervised transcripts for 
training \cite{lamel2000lightly}.  

\begin{figure}[t!]
	\centering
	\includegraphics[width=0.46\textwidth,height=0.31\textwidth]{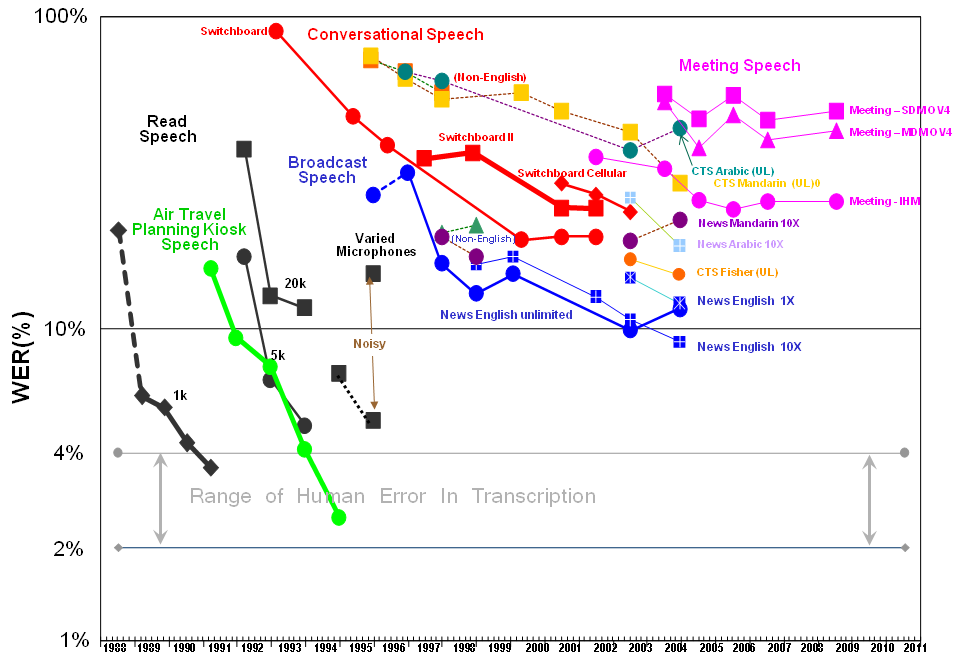}
	\caption{The NIST STT Benchmark Test History - May'09 \cite{nist}}
	\label{fig:results}
	\vspace{-10pt}
\end{figure}

The EARS program led to significant advances in speech recognition technology for both 
domains, with the development of techniques that could ingest large quantities of unsupervised or 
semi-supervised training data, discriminative training of recognition models, methods to deal with channel and speaker 
variabilities in the data, real-time decoding of test data, and also approaches to combine outputs from various systems \cite{gales2006progress,matsoukas2006advances,chen2006advances,stolcke2006recent}. 
Several of these techniques have further been extended to build ASR systems on broadcast news data in 
various languages \cite{sinha2006cu,lei2006improved,xiang2006morphological}. Figure 1 shows progress made in this domain over the past two decades.
More recently, as part of the MGB Challenge, in addition to the core ASR problem, several other related tasks - speaker 
diarization and lightly supervised alignment of data have also been studied \cite{bell2015mgb,ali2017speech}.

In  \cite{saon2017english,kurata2017language} we describe state-of-the-art speech recognition systems on the CTS task using multiple LSTM and 
ResNet acoustic models trained on various acoustic features along with word and character LSTMs 
and convolutional WaveNet-style language models. This advanced recipe achieves 5.1\% and 9.9\% on the Switchboard 
and CallHome subsets of the Hub5 2000 evaluation. In this paper we develop a similar but simpler variant for BN. As described earlier, by developing this system we investigate how these earlier
 proposed systems can be trained on BN data which are not human annotated but are created using closed captions.
 To create these systems, instead of adding all the available training data we carefully select a reliable subset.
 We then train LSTM and residual network based acoustic models with a combination of n-gram and neural network 
 language models on this selected data. In addition to automatic speech recognition results, similar to  \cite{saon2017english}, we also 
 present human performance on the same BN test sets. These evaluations allow us to properly benchmark our 
 automatic system performance. Similar to earlier human performance evaluations on CTS, we observe a significant 
 gap between human and automatic results.

The rest of the paper is organized as follows. In Section 2 we describe the human evaluation experiments on
two broadcast news test sets - RT04 and DEV04F. We also compare the recognition errors we observe 
with human and automatic recognition systems. Section 3 describes the development of our ASR systems - training
data selection, acoustic and language model building. In Section 4 we present WER results using the proposed
system. The paper concludes with a discussion in Section 5.

\section{Human Transcription Experiments}
\label{sec:human}

Similar to \cite{saon2017english}, human performance measurements on two broadcast news tasks - RT04 and DEV04F - are carried out by Appen. 
For these evaluations we limit the audio from the test sets to only regions of speech that are marked for scoring using the
original references and scoring scripts provided during the EARS evaluation. After processing, the 
RT04 test set has 4 hours of BN data from 12 shows with about 230 overlapping speakers across the shows. 
The DEV04F test set is smaller, with about 2 hours of data from 6 shows with close to 100 overlapping speakers 
across the shows. 

The first round of transcripts was produced by three independent transcribers, followed by  
quality checking by a fourth senior transcriber. All four transcribers are native US English speakers 
and were selected based on the quality of their work on past transcription projects. The transcriptions 
were produced in line with LDC transcription guidelines for hyphenations, spelled abbreviations, contractions, 
partial words, non-speech sounds, etc. that were used to produce the original transcripts for these test sets.
The three primary transcribers took 14-16 times real-time (xRT) for the first pass followed by
 an additional 3xRT for the second quality checking pass (by Transcriber 4). Both passes 
involved listening to the audio multiple times: around 3-4 times for the first pass and 1-2 times for the
second. In order to use NIST’s scoring tool, sclite \cite{nisttools}, the human annotations were converted into 
 CTM files which have time-marked word boundary information. The transcriptions were also filtered to
 remove non-speech markers, partial words, punctuation marks etc as described in \cite{saon2017english}.
 Table 1 shows the error rates  of the three transcribers after quality checking by the fourth transcriber.

\begin{table}[h!]
	\begin{tabular}{|c|c|c|}
		\hline
		        & DEV04F & RT04 \\
		\hline
		\hline
		Transcriber 1 & 4.4  & 3.6 \\
		\hline
		Transcriber 2 & 4.4 & 3.2 \\
		\hline
		Transcriber 3 & \textbf{3.6} & \textbf{2.8} \\
		\hline
	\end{tabular}
     \centering
     \caption{Human Performance (WER\%) on RT04 and DEV04F.}
	\label{tab:hp}
\end{table}

Compared to the human transcription results on the CTS tasks, 5.1\% and 6.8\% on the Switchboard and CallHome subsets 
of the Hub5 2000 evaluation \cite{saon2017english}, the word error rate on BN is much lower. Although this reduction could be because 
BN speech is well articulated, the transcribers reported that these test sets were 
much denser with respect to speech content, had considerable background noise,
and a significant number of named entities that required lookup to ensure correctness, compared to traditional CTS test sets. 
The best WER results we obtain, 3.6\% and 2.8\%, also fit in the expected human transcription error range indicated in Figure \ref{fig:results}.
A more detailed error analysis and comparison of human and automatic recognition is presented in the Discussion section.

\section{ASR System Building}

As described earlier, one differentiating characteristic of ASR system builds for this BN task is the limited amount
of carefully annotated manual transcriptions. Prior to the EARS program, LDC released about 144 hours of 
careful manual annotations for a portion of the Hub4 acoustic training data collected between May 1996
and January 1998. In addition to this, several sources of BN data were available for training acoustic
models during the EARS program period with just closed caption transcripts. These data sources include about 
1000 hours of data as part of different data releases collected between 1998-2001 (TDT2 and TDT4) and 
about 7000 hours of broadcast news released in 2003 as part of the EARS program (BN03). In this paper
we use processed versions of these data sources to build deep neural network based acoustic and language models.

\subsection{Training Data Preparation}
\label{sec:data}

To process the BN data with noisy closed captions, the data is first decoded
with multiple off-the-shelf broadband ASR systems using a biased LM created with the available closed captions.
Based on the confidence scores of these decodes and agreement between the multiple system decodes, we perform
a strict filtering of the data to create 2 sub-corpora that we consider have very reliable transcripts -
\begin{itemize}
	\item The BN-400 Corpus - This is a corpus of about 430 hours of broadcast news data selected from the data sources
	described above. This data corpus includes 144 hours of carefully transcribed audio along with data with semi-supervised 
	transcripts created via a biased decode of the matching audio.
	\item The BN-1300 Corpus - This corpus is an extended version of the BN-400 corpus with about 900 additional hours of broadcast news.
\end{itemize}

\subsection{Acoustic Model Development}
\label{sec:am}

As discussed earlier, one of the key objectives of this work is to verify the usefulness of our earlier proposed
system strategy for CTS. In \cite{saon2017english}, two kinds of acoustic models, a convolutional and a non-convolutional acoustic model
with comparable performance, are used since they produce good complementary outputs which can be further combined
for improved performance. The convolutional network used in that work is a residual network (ResNet) and an LSTM is
used as the non-convolutional network. The acoustic scores of these systems are subsequently combined for the final decodes.
Similar to that work, in this paper also we train ResNet and LSTM based acoustic models.

Both these acoustic models are based on speaker transformed features. The ResNet uses 40 dimensional VTL-warped
log-mel features along with their $\Delta$ and $\Delta\Delta$ transforms. The LSTM based network is trained on
40 dimensional FMLLR features appended with 100 dimensional ivectors and 40 dimensional VTL-warped
log-mels along with their $\Delta$ and $\Delta\Delta$ parameters. The speaker transformed features - 
FMLLR and VTL-warped features and ivectors, are derived using traditional HMM-GMM based systems trained 
on the BN-400 corpus. We model the acoustic space with 32K context-dependent HMM states.

%\subsubsection{LSTM acoustic model}

Similar to the architecture in \cite{saon2017english}, we train an LSTM acoustic model with 6 bidirectional layers having
1024 cells per layer (512 per direction), one linear bottleneck layer with 256 units and an output layer 
with 32K units corresponding to the context-dependent HMM states we derived in the HMM-GMM  system build. 
The model is trained using non-overlapping subsequences of 21 frames. Subsequences from different utterances are grouped into 
mini-batches of size 128 for processing speed and reliable gradient estimates. After the cross-entropy 
based training on the  BN-1300 Corpus has converged we also sequence train the model using the 144 hours of 
carefully transcribed audio. 

%\subsubsection{ResNet acoustic model}

To complement the LSTM acoustic model, we train a deep Residual Network based on the best
performing architecture proposed in \cite{saon2017english}. The ResNet has 12 residual blocks  
followed by 5 fully connected layers. To effectively train this network with 25 convolutional layers, 
a short-cut connection is placed between each residual block to allow for additional flow of information 
and gradients. Each layer has a batch normalization layer as well. 
Table 2 gives a summary of the network architecture of the ResNet model.
The ResNet consists of several stages with different numbers of feature maps in each stage: 
64 in stage 1, 128 in stage 2, 256 in stage 3 and 512 in stage 4.
Each stage has an “initStride”  which indicates the (frequency $\times$ time) stride for the first block of
that stage as the number of feature maps is increased. The stride applies to both the first 3$\times$3 
convolution of each block and also the 1$\times$1 convolution in projection shortcut between each block.
The ResNet acoustic model is trained using the cross-entropy training criterion on the BN-1300 Corpus and 
then sequence trained using the 144 hours of carefully transcribed audio.

\begin{table}[t!]
	\resizebox{0.5\textwidth}{!}{%
		\begin{tabular}{|l | l|}
			\hline
			Input & 3 $\times$ 40 $\times$ 76\\
			\hline
			&  conv 5x5, 64; \\
			Stage 0 &  maxpool (2$\times$1) \\
			\hline
			&  initStride 1 $\times$ 1; \\ 
			Stage 1 &  3 $\times$ [conv 3$\times$3, 64 feat. maps, conv 3$\times$3, 64 feat. maps] \\
			\hline	
			&  initStride 2 $\times$ 1; \\ 
			Stage 2 &  3 $\times$ [conv 3$\times$3, 128 feat. maps, conv 3$\times$3, 128 feat. maps] \\
			\hline
			&  initStride 2 $\times$ 1; \\ 
			Stage 3 &  3 $\times$ [conv 3$\times$3, 256 feat. maps, conv 3$\times$3, 256 feat. maps] \\
			\hline
			&  initStride 2 $\times$ 2; \\ 
			&  3 $\times$ [conv 3$\times$3, 512 feat. maps, conv 3$\times$3, 512 feat. maps]; \\
			Stage 4 & maxpool (2$\times$2) \\
			\hline
			Output & 3 $\times$ FC 2084; FC 1024; FC 32K \\
			\hline
	\end{tabular}}
	\centering
	\caption{ResNet acoustic model architecture}
	\label{tab:resnet}
\end{table}

\subsection{Language Model Development}
\label{sec:lm}

Similar to the development of acoustic models, several kinds of n-gram and neural network
based language models are built on this BN task.  For the initial decode that produces word lattices, 
an n-gram and a feed forward neural network language model are first built. To rescore the word lattices 
and n-best lists produced by these models, advanced LSTM based NN language models are also constructed.

%\subsubsection{N-gram language model}
%\subsubsection{Feed forward neural network model}
%\subsubsection{LSTM neural network model}

The primary language model training text for all these models consists of a total of 350M words 
from different publicly available sources released by LDC during the GALE \cite{olive2011handbook} and EARS 
evaluation periods suitable for broadcast news. The baseline language model is a linear 
interpolation of word 6-gram models, one for each corpus with a vocabulary size of about 80K words.
We train a feed forward neural network model based on the same data and vocabulary as the n-gram
language model described above. The neural network model (FFNN-LM) uses an embedding size of 120, 
a hidden layer size of 1200 and the maxout non-linearity \cite{yinghui2018fast}. We use noise contrastive estimation
to train this unnormalized NNLM \cite{sethy2015}. For decoding experiments, the FF-NNLM is interpolated with the baseline 
6-gram arpabo with an interpolation weight set to 0.5.

\begin{table}[t!]
	\begin{tabular}{|l|l|c|c|}
		\hline
		AM & LM  & DEV04F & RT04 \\
		\hline
		\hline
		LSTM & n-gram  & 7.6 & 7.7\\
		\hline
		ResNet & n-gram  & 9.6 & 8.9\\
		\hline
		\hline
		LSTM & n-gram + FFNN-LM  & 7.2 & 7.0  \\
		\hline
		ResNet & n-gram + FFNN-LM & 9.0 & 8.1 \\
		\hline
		\hline
		LSTM+ResNet & n-gram + FFNN-LM & 7.2 & 7.0 \\
		\hline
	\end{tabular}
	\centering
	\caption{ASR decoding results (WER\%) on RT04 and DEV04F.}
	\label{tab:ngram}
\end{table}

\begin{table}[t!]
	\begin{tabular}{|c|c|c|}
		\hline
		& DEV04F & RT04 \\
		\hline
		\hline
		LSTM1-LM rescoring & 6.6 & 6.1 \\
		\hline
		LSTM2-LM rescoring & 6.6 & 6.1 \\
		\hline
		LSTM1/LSTM2-LM rescoring & 6.5 & 5.9 \\
		\hline
	\end{tabular}
	\centering
	\caption{LSTM rescoring results (WER\%) on RT04 and DEV04F.}
	\label{tab:lstm}
\end{table}
In addition to the n-gram and feed forward neural network language models, we also train two different 
flavors of LSTM language models with the same vocabulary and training data as described above. 
The first LSTM model (LSTM1-LM) consists of one word embedding layer with 256 units, four LSTM layers with 1024 
units, one fully-connected layer, and one softmax layer. The second to fourth LSTM layers and the 
fully-connected layer allow residual connections \cite{he2016deep}. Dropout is applied to the vertical dimension 
only and not applied to the time dimension. We trained this model to minimize the  cross-entropy 
objective using Adam for learning rate control \cite{kingma2014adam}. The second LSTM based LM (LSTM2-LM) consists of two LSTM layers, 
 each layer with 2048 nodes and a word embedding size of 512. Before the softmax-based estimation of an 80K-dimensional posterior vector, the feature space was reduced to 128 by a linear bottleneck layer.
During the training various dropout techniques were applied \cite{merity2018regularizing}. First, the outputs of the embedding 
and each LSTM layer were masked at a 10\% rate. Second, 10\% dropout was also applied on the embedding weights, 
and also on the parameters of the recurrent connection of the LSTMs. These weight masks were kept constant 
during processing a mini-batch of sequences. In the final step of the training, the model was fine-tuned on 
the best matching resource, the EARS BN data. The SGD based model training uses a batch size of 128 
and a Nesterov momentum of 0.9 to optimize model parameters on the cross-entropy criterion.

\section{ASR Experiments and Results}

The acoustic and language models described above are used to decode the  RT04 and DEV04F test sets.
We use the same speech segments that were provided to the human annotators for our various experiments.
In our first set of experiments we separately test the LSTM and ResNet models in conjunction with the
n-gram and FF-NNLM, before combining scores from the two acoustic models. Table 3 shows the individual 
and combined results we obtain on both the test sets. In comparison with the results
obtained on the CTS evaluation with similar acoustic models \cite{saon2017english}, the LSTM and ResNet operate
at similar WERs. Unlike results observed on the CTS task, no significant reduction in WER is  
obtained after scores from both the LSTM and ResNet models are combined. The LSTM model with an n-gram LM  
individually performs quite well and its results further improve with the addition of the FF-NNLM.

For our second set of experiments word lattices are generated after decoding with the 
LSTM+ResNet+n-gram+FF-NNLM model. We generated n-best lists from these lattices and rescored them with the LSTM1-LM.
LSTM2-LM is also used to rescore word lattices independently. Table 4 shows the results after our rescoring experiments.
We observe significant WER gains after using the LSTM LMs similar to those reported in \cite{saon2017english}. 
%Although both the LSTM LMs perform at the same WER, they produce complimentary outputs that improve performance further
%when combined.
By rescoring outputs with both LSTM1 and LSTM2,  we achieve new performance milestones with final WERs of 6.5\% and 5.9\% on DEV04F and RT04 respectively.

Our ASR results have clearly improved state-of-the-art performance on these test sets compared to the various results
reported in \cite{gales2006progress, matsoukas2006advances, sainath2013improvements}. Significant progress has also been made 
compared to systems developed over the last decade, as shown in Figure \ref{fig:results}.
\begin{table}[t!]
	\centering
	\resizebox{0.7\columnwidth}{!}{
		\begin{tabular}{|l|l|l|l|l|} \hline
			&	\multicolumn{2}{|c|}{DEV04F} & \multicolumn{2}{c|}{RT04}\\ \hline
			&	ASR              & Human            & ASR & Human \\ \hline
			Sub &	3.2     & 1.9  & 3.1    & 1.6 \\ \hline
			Del &   2.2     & 0.8  & 2.2    & 0.6 \\ \hline
			Ins &   1.1     & 0.9  & 0.6    & 0.6 \\ \hline
			All &   6.5     & 3.6  & 5.9    & 2.8 \\ \hline
	\end{tabular}}
	\caption{\label{over} Overall substitution, deletion and insertion errors of humans and ASR system.}
\end{table}

\section{Discussions}

When compared to the human performance results, the absolute ASR WER is about 3\% worse.  From Table 5 we observe
that although the machine and human insertion error rates are comparable, the ASR system has much higher substitution 
and deletion error rates. Tables 6 and 7 list the 10 most frequent errors of each type.
We draw the following observations based on these errors -
\begin{enumerate}
\item There is a significant overlap in the words that ASR and humans delete, substitute and insert.
\item  Humans seem to be careful about marking hesitations - \textit{\%hesitation} is the most inserted symbol. Hesitations seem to be important in conveying meaning to the sentences in human transcriptions. The ASR systems however focus on blind recognition and not in improving the meaning with appropriate pauses, etc. To measure the extent of this process, we score the human transcripts without any hesitations in a separate experiment and observe a 0.1\% absolute improvement in WER.
\item  Machines have trouble recognizing short function words - \textit{\{the, and, of, a, that\}} and these get deleted the most. Humans, 
on the other hand, seem to catch most of them. It is likely that these words are not fully articulated and hence the machine fails to recognize them  
while humans are able to infer these words even without full acoustic evidence since they may have a better model of syntax/semantics.
\item  Compared to the telephone conversation confusions recorded in  \cite{saon2017english} - one symbol that is clearly missing is the back-channel response - this is probably from the very nature of the BN domain.
\item Similar to telephone conversation confusions reported in  \cite{saon2017english}, humans performance is much higher because the number of deletions is significantly lower - compare
2.3\% vs 0.8\%/0.6\% for deletion errors in Table 5.
\end{enumerate}

\begin{table}[t!]
	\centering
	\resizebox{\columnwidth}{!}{
		\begin{tabular}{|l|l|l|l|} \hline
			\multicolumn{2}{|c|}{DEV04F} & \multicolumn{2}{c|}{RT04}\\ \hline
			ASR              & Human            & ASR & Human \\ \hline
			8: and / in      & 5: too / to      & 21: the / a     & 19: the / a \\
			7: had / have    & 4: is / has      & 16: and / in    & 15:  and / in \\
			3: a / the      & 4: a / the       & 15: a / the     & 11:  in / and \\
			3: has / is       & 4: had / have    & 14: has / is    & 7:  (\%hes) / a \\
			3: on / in       & 3: the / (\%hes) & 11: (\%hes) / a & 6:  this / the \\
			3: that / it    & 3: on / in       & 11: in / and    & 5:  are / were\\
			3: too / to   & 3: (\%hes) / a   & 8: that / it    & 4:  and / then \\
			3: this / the      & 3: in / and      & 7: this / the   & 4:  as / is \\
			2: (\%hes) / and        & 3: and / in      & 6: is / as      & 3:  (\%hes) / and \\
			2: and / an      & 2: are / were    & 4: it / that    & 2: an / in \\ \hline
	\end{tabular}}
	\caption{\label{subs}Most frequent substitution errors for humans and ASR systems on DEV04F and RT04}
\end{table}

\begin{table}[t!]
	\centering
	\resizebox{\columnwidth}{!}{
		\begin{tabular}{|l|l|l|l||l|l|l|l|} \hline
			\multicolumn{4}{|c||}{Deletions} & \multicolumn{4}{c|}{Insertions}\\ \hline
			\multicolumn{2}{|c|}{DEV04F}   & \multicolumn{2}{c||}{RT04} & \multicolumn{2}{c|}{DEV04F} & \multicolumn{2}{c|}{RT04}\\ \hline
			ASR      & Human   & ASR       & Human     & ASR        & Human       & ASR      & Human\\ \hline
			49: the  & 13: the &  92: and  & 21: and   & 15: and    & 33: (\%hes) & 10: i    & 31: (\%hes)\\
			43: and  & 10: a   &  69: the  & 21: the   & 12: colon  & 14: and     & 8: and   & 13: a\\
			21: a    & 8: and  &  52: a    & 17: a     & 10: to     & 10: the     & 7: it    & 11: and\\
			17: that & 8: that &  47: it   & 14: is    & 8:  call   & 7: it       & 6: that  & 10: c\\
			17: to   & 6: it   &  35: that & 9: in     & 8:  the    & 5: is       & 6: to    & 9: have\\
			16: it   & 5: i    &  33: is   & 9: that   & 7:  a      & 4: a        & 4: a     & 8: the\\
			16: of   & 5: of   &  32: in   & 8: are    & 7:  ask    & 4: that     & 4: post  & 6: it\\
			15: you  & 5: you  &  28: you  & 8: i      & 5:  are    & 4: to       & 3: are   & 5: post\\
			14: are  & 4: are  &  22: of   & 8: of     & 4:  how    & 3: are      & 3: had   & 5: to\\
			10: in   & 3: have &  19: i    & 7: (\%hes)& 3:  be     & 3: c        & 3: he    & 4: are\\ \hline
	\end{tabular}}
	\caption{\label{del-ins}Most frequent deletion and insertion errors for humans and ASR systems on DEV04F and RT04}
\end{table}

\section{Conclusion}
\label{sec:conclusions}

We have presented recent improvements on broadcast news transcription based on earlier
established techniques shown to be useful on CTS. Our experiments on BN show that these techniques
can be transferred across domains to provide highly accurate transcriptions. For
both acoustic and language modeling we have demonstrated the effectiveness of LSTM and ResNet
 based models. To verify the extent of the improvements
obtained, human evaluation experiments are also performed on the two test sets
of interest. We show that there still exists a significant gap between human and
machine performance and demonstrate the need for continued research on broadcast news.

\vspace*{\fill}
\pagebreak
% References should be produced using the bibtex program from suitable
% BiBTeX files (here: strings, refs, manuals). The IEEEbib.bst bibliography
% style file from IEEE produces unsorted bibliography list.
% -------------------------------------------------------------------------
\bibliographystyle{IEEEbib}
\bibliography{refs}

\begin{thebibliography}{10}

\bibitem{olive2011handbook}
Joseph Olive, Caitlin Christianson, and John McCary,
\newblock {\em Handbook of natural language processing and machine translation:
  {DARPA} {G}lobal {A}utonomous {L}anguage {E}xploitation},
\newblock Springer Science \& Business Media, 2011.

\bibitem{saon2017english}
George Saon, Gakuto Kurata, Tom Sercu, Kartik Audhkhasi, Samuel Thomas,
  Dimitrios Dimitriadis, Xiaodong Cui, Bhuvana Ramabhadran, Michael Picheny,
  Lynn-Li Lim, et~al.,
\newblock ``English conversational telephone speech recognition by humans and
  machines,''
\newblock {\em arXiv preprint arXiv:1703.02136}, 2017.

\bibitem{kurata2017language}
Gakuto Kurata, Bhuvana Ramabhadran, George Saon, and Abhinav Sethy,
\newblock ``Language modeling with highway {LSTM},''
\newblock in {\em Automatic Speech Recognition and Understanding Workshop
  (ASRU), 2017 IEEE}. IEEE, 2017, pp. 244--251.

\bibitem{xiong2018microsoft}
Wayne Xiong, Lingfeng Wu, Fil Alleva, Jasha Droppo, Xuedong Huang, and Andreas
  Stolcke,
\newblock ``The {M}icrosoft 2017 conversational speech recognition system,''
\newblock in {\em 2018 IEEE International Conference on Acoustics, Speech and
  Signal Processing (ICASSP)}. IEEE, 2018, pp. 5934--5938.

\bibitem{han2017capio}
Kyu~J Han, Akshay Chandrashekaran, Jungsuk Kim, and Ian Lane,
\newblock ``The {CAPIO} 2017 conversational speech recognition system,''
\newblock {\em arXiv preprint arXiv:1801.00059}, 2017.

\bibitem{lamel2000lightly}
Lori Lamel, Jean-Luc Gauvain, and Gilles Adda,
\newblock ``Lightly supervised acoustic model training,''
\newblock in {\em ASR2000-Automatic Speech Recognition: Challenges for the new
  Millenium ISCA Tutorial and Research Workshop (ITRW)}, 2000.

\bibitem{nist}
NIST,
\newblock ``{NIST Rich Transcription Evaluation},''
  \url{https://www.nist.gov/itl/iad/mig/rich-transcription-evaluation}.

\bibitem{gales2006progress}
Mark~JF Gales, PC~Woodland, Ho~Yin Chan, David Mrva, Rohit Sinha, Sue~E
  Tranter, et~al.,
\newblock ``Progress in the {CU-HTK} broadcast news transcription system,''
\newblock {\em IEEE Transactions on Audio, Speech, and Language Processing},
  vol. 14, no. 5, pp. 1513--1525, 2006.

\bibitem{matsoukas2006advances}
Spyridon Matsoukas, J-L Gauvain, Gilles Adda, Thomas Colthurst, Chia-Lin Kao,
  Owen Kimball, Lori Lamel, Fabrice Lefevre, Jeff~Z Ma, John Makhoul, et~al.,
\newblock ``Advances in transcription of broadcast news and conversational
  telephone speech within the combined {EARS} {BBN/LIMSI} system,''
\newblock {\em IEEE Transactions on Audio, Speech, and Language Processing},
  vol. 14, no. 5, pp. 1541--1556, 2006.

\bibitem{chen2006advances}
Stanley~F Chen, Brian Kingsbury, Lidia Mangu, Daniel Povey, George Saon, Hagen
  Soltau, and Geoffrey Zweig,
\newblock ``Advances in speech transcription at {IBM} under the {DARPA EARS
  program},''
\newblock {\em IEEE Transactions on Audio, Speech, and Language Processing},
  vol. 14, no. 5, pp. 1596--1608, 2006.

\bibitem{stolcke2006recent}
Andreas Stolcke, Barry Chen, Horacio Franco, Venkata Ramana~Rao Gadde, Martin
  Graciarena, Mei-Yuh Hwang, Katrin Kirchhoff, Arindam Mandal, Nelson Morgan,
  Xin Lei, et~al.,
\newblock ``Recent innovations in speech-to-text transcription at
  {SRI-ICSI-UW},''
\newblock {\em IEEE Transactions on Audio, Speech, and Language Processing},
  vol. 14, no. 5, pp. 1729--1744, 2006.

\bibitem{sinha2006cu}
Rohit Sinha, Mark~JF Gales, DY~Kim, X~Andrew Liu, Khe~Chai Sim, and Philip~C
  Woodland,
\newblock ``The {CU-HTK} {M}andarin broadcast news transcription system,''
\newblock in {\em Acoustics, Speech and Signal Processing, 2006. ICASSP 2006
  Proceedings. 2006 IEEE International Conference on}. IEEE, 2006, vol.~1, pp.
  I--I.

\bibitem{lei2006improved}
Xin Lei, Manhung Siu, Mei-Yuh Hwang, Mari Ostendorf, and Tan Lee,
\newblock ``Improved tone modeling for {M}andarin broadcast news speech
  recognition,''
\newblock in {\em Ninth International Conference on Spoken Language
  Processing}, 2006.

\bibitem{xiang2006morphological}
Bing Xiang, Kham Nguyen, Long Nguyen, Richard Schwartz, and John Makhoul,
\newblock ``Morphological decomposition for {A}rabic broadcast news
  transcription,''
\newblock in {\em Acoustics, Speech and Signal Processing, 2006. ICASSP 2006
  Proceedings. 2006 IEEE International Conference on}. IEEE, 2006, vol.~1, pp.
  I--I.

\bibitem{bell2015mgb}
Peter Bell, Mark~JF Gales, Thomas Hain, Jonathan Kilgour, Pierre Lanchantin,
  Xunying Liu, Andrew McParland, Steve Renals, Oscar Saz, Mirjam Wester,
  et~al.,
\newblock ``The {MGB} challenge: Evaluating multi-genre broadcast media
  recognition,''
\newblock in {\em Automatic Speech Recognition and Understanding (ASRU), 2015
  IEEE Workshop on}. IEEE, 2015, pp. 687--693.

\bibitem{ali2017speech}
Ahmed Ali, Stephan Vogel, and Steve Renals,
\newblock ``Speech recognition challenge in the wild: {Arabic MGB-3},''
\newblock in {\em Automatic Speech Recognition and Understanding Workshop
  (ASRU), 2017 IEEE}. IEEE, 2017, pp. 316--322.

\bibitem{nisttools}
NIST,
\newblock ``{NIST Tools},'' \url{https://www.nist.gov/itl/iad/mig/tools}.

\bibitem{yinghui2018fast}
Yinghui Huang, Abhinav Sethy, and Bhuvana Ramabhadran,
\newblock ``Fast neural network language model lookups at n-gram speeds,''
\newblock in {\em Interspeech 2017}. ISCA, 2017.

\bibitem{sethy2015}
Abhinav Sethy, Stanley Chen, Ebru Arisoy, and Bhuvana Ramabhadran,
\newblock ``Unnormalized exponential and neural network language models,''
\newblock in {\em ICASSP}, 2015.

\bibitem{he2016deep}
Kaiming He, Xiangyu Zhang, Shaoqing Ren, and Jian Sun,
\newblock ``Deep residual learning for image recognition,''
\newblock in {\em Proceedings of the IEEE conference on computer vision and
  pattern recognition}, 2016, pp. 770--778.

\bibitem{kingma2014adam}
Diederik~P Kingma and Jimmy Ba,
\newblock ``Adam: A method for stochastic optimization,''
\newblock {\em arXiv preprint arXiv:1412.6980}, 2014.

\bibitem{merity2018regularizing}
Stephen Merity, Nitish~Shirish Keskar, and Richard Socher,
\newblock ``Regularizing and optimizing {LSTM} language models,''
\newblock in {\em International Conference on Learning Representations (ICLR)},
  2018.

\bibitem{sainath2013improvements}
Tara~N Sainath, Brian Kingsbury, Abdel-rahman Mohamed, George~E Dahl, George
  Saon, Hagen Soltau, Tomas Beran, Aleksandr~Y Aravkin, and Bhuvana
  Ramabhadran,
\newblock ``Improvements to deep convolutional neural networks for {LVCSR},''
\newblock in {\em Automatic Speech Recognition and Understanding (ASRU), 2013
  IEEE Workshop on}. IEEE, 2013, pp. 315--320.

\end{thebibliography}

\end{document}